\pdfoutput=1

\documentclass[11pt]{article}

\usepackage[preprint]{acl}

\usepackage{times}
\usepackage{latexsym}

\usepackage[T1]{fontenc}

\usepackage[utf8]{inputenc}

\usepackage{microtype}

\usepackage{inconsolata}

\usepackage{microtype}
\usepackage{graphicx}
\usepackage{subfigure}
\usepackage{booktabs} 
\usepackage{colortbl}

\usepackage{amsmath}
\usepackage{amssymb}
\usepackage{mathtools}
\usepackage{amsthm}
\usepackage{multirow}
\usepackage{enumitem}

\usepackage{algorithm}
\usepackage{multirow}
\usepackage[noend]{algorithmic}
\usepackage{wrapfig}
\usepackage{framed}

\definecolor{babyblue}{HTML}{4285f4}
\definecolor{LightGrey}{HTML}{d9d9d9}

\newcommand{\method}{{\textbf{RefreshKV}}~}
\newcommand{\methodN}{{RefreshKV}}
\newcommand{\methodE}{{\textbf{RefreshKV}}}

\newcommand{\qwen}{{Qwen2-7B}}
\newcommand{\llama}{{Llama-3.1-8B}}

\title{RefreshKV: Updating Small KV Cache During Long-form Generation }

\author{Fangyuan Xu$^{1}$, Tanya Goyal$^{2*}$, Eunsol Choi$^{1}$\thanks{Equal advising.}\\
Department of Computer Science\\
$^1$New York University,
$^2$Cornell University \\
\texttt{\{fx2145,eunsol\}@nyu.edu}, \texttt{tanyagoyal@cornell.edu}
}

\begin{document}
\maketitle
\begin{abstract}
Generating long sequences of tokens given a long-context input is a very compute-intensive inference scenario for large language models (LLMs). One prominent inference speed-up approach is to construct a smaller key-value (KV) cache, relieving LLMs from computing attention over a long sequence of tokens. While such methods work well to generate short sequences, their performance degrades rapidly for long-form generation. Most KV compression happens once, prematurely removing tokens that can be useful later in the generation. We propose a new inference method, \methodE, that flexibly alternates between full context attention and attention over a subset of input tokens during generation. After each full attention step, we update the smaller KV cache based on the attention pattern over the entire input. Applying our method to off-the-shelf LLMs achieves comparable speedup to eviction-based methods while improving performance for various long-form generation tasks. Lastly, we show that continued pretraining with our inference setting brings further gains in performance. 
\end{abstract}

\section{Introduction}

Large language models (LLMs) are capable of ingesting extremely long inputs and generating long outputs \citep{llama3,team2024google}. Yet, deploying such long-context LLMs is very costly. As the context length increases, 
memory usage for storing the key-value (KV) cache increases linearly, while attention computation scales quadratically. These two factors lead to high latency during inference; \citet{adnan2024keyformer} reports 50x latency increase as context length increased 16x 
for the MPT-7B model~\citep{MosaicML2023Introducing}. 

Prior works~\citep{Beltagy2020Longformer,Child2019GeneratingLS,Xiao2023EfficientSL,zhang2024h2o,li2024snapkv,adnan2024keyformer} propose to maintain a smaller KV cache by evicting a subset of past tokens. These approaches improve both the memory and computation efficiency, as the KV cache of only a subset of tokens will be kept and attention computation is reduced. However, once an input token is eliminated from the KV cache (either based on locality assumption~\citep{Xiao2023EfficientSL} or by eviction during the generation process~\citep{zhang2024h2o}), one cannot recover eliminated tokens. We find that while such methods show minor degradation compared to full KV cache in short-form generation tasks, their performance degrades rapidly for long-form generation tasks. 


\begin{figure*}[t]
    \centering\vspace{-1em}
    \includegraphics[scale=0.23,trim=20mm 0mm 0mm 0mm]{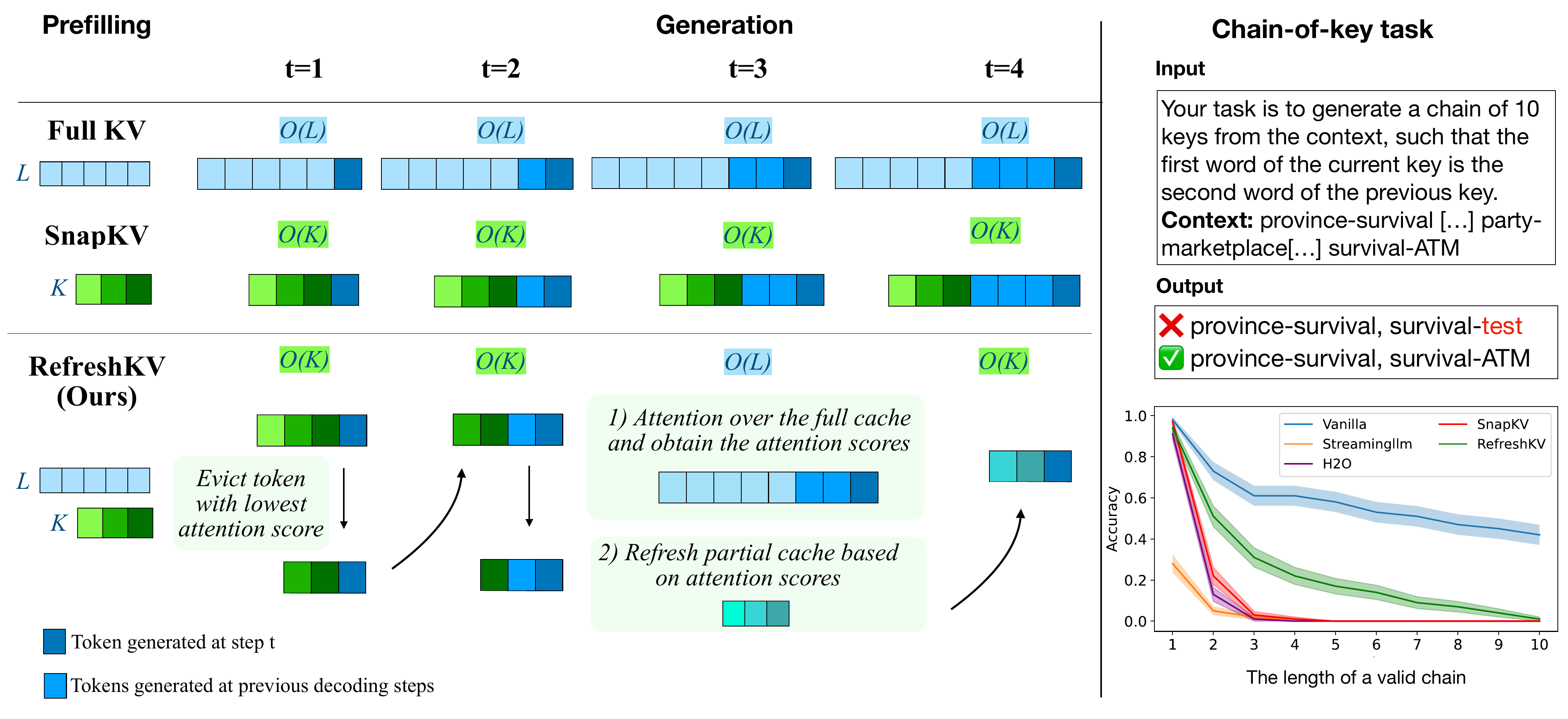}
    \caption{Left: Illustration of \method (with $L=5$, $K=3$ and a stride $S=3$) compared to baseline (SnapKV and Full KV) when generating four tokens. The figure shows the computation complexity of attention operation, and the size of the KV cache used at each decoding step for each method. Our approach alternates between inferencing with the partial cache(t=1,2,4) and the full cache(t=3). Compared to eviction-based method (e.g. SnapKV) which completely discard the evicted tokens, \method updates the partial cache based on attention scores over the entire context during the full attention steps. Right: An example of the chain-of-key task and performance of \method and the baselines. \methodN~maintains performances across different length while eviction-based baeslines' performance degrades when generating a chain with more than one key.} \vspace{-1em} 
    \label{fig:RecycledAttention-Figure1}
\end{figure*}


Having observed the limitations of existing approaches, we propose a novel approach, \methodE, which periodically refreshes the smaller KV cache during the generation process. Our method keeps the full KV cache throughout inference (thus no gain in memory footprint), but perform attention over a dynamically constructed small KV cache to achieve inference speedups. Our method alternates between two modes of generation: generation that attends over the full KV cache and generation that attends over a smaller KV cache with subset of tokens (see Figure~\ref{fig:RecycledAttention-Figure1}). To construct the smaller KV, we identify the topK attended tokens from the most recent step that attends over the full KV cache, observing that consecutive tokens have similar attention pattern~\citep{li2024snapkv}.

A key component of \methodN~is deciding when to perform the computationally expensive full attention steps and refresh the small KV cache. Instead of mandating a fixed (and potentially suboptimal) schedule, \methodN~compares the query embedding similarity of the current and previous full attention step, and dynamically triggers full attention step when the similarity is low. 
Our approach (no KV eviction, dynamically constructed smaller KV, low latency) establishes a middle ground between full attention (no KV eviction, high latency, high performance) and sparse attention (KV eviction, reduced latency, low performance), particularly useful for long-form generation.

Our method can be applied to any off-the-shelf LLM. We experiment with two long-context LLMs, \llama{} \citep{llama3} and \qwen{} \citep{Yang2024Qwen2TR}. We compare against KV eviction baselines StreamingLLM \cite{Xiao2023EfficientSL}, H$_2$O \cite{zhang2024h2o} and SnapKV \cite{li2024snapkv} on the long-range language modeling task and a suite of downstream long-context tasks \cite{bai2023longbench, zhang-etal-2024-bench, ye25longproc} that require long outputs given long inputs.

Our experiments show that \method outperforms eviction-based methods in both these settings, with similar level of speed-up. In particular, we examine two long-form generation tasks that are not evaluated by previously proposed eviction-based methods: (1) when majority of tokens are required to generate the output (e.g. converting information in an HTML page to a TSV file) and (2) when the important tokens required at the current generation step is dependent on the previously generated tokens (a new task, \textbf{Chain-of-key}, as depicted in Figure \ref{fig:RecycledAttention-Figure1}). While eviction-based methods such as H$_2$O and SnapKV fail completely in HTML to TSV task \citep{ye25longproc}, achieving 0 F1 score, \methodN~ recovers 52\% of the performance. 
Our analysis shows that the performance gains are attributed to updating the  partial cache rather than occasionally attending to the entire output.
Lastly, we explore continued pretraining \llama{} with \methodN, which leads to further improvements. Our contributions are as follows:\footnote{We will release our code at \url{https://github.com/carriex/refreshkv}.}

\begin{itemize}[leftmargin=*]
    \item We identify the failures of existing KV eviction methods when LLMs are tasked with challenging long-form generation.
    \item Motivated by the failures of KV cache eviction methods, we introduce a new inference method, \methodE, that rebuilds a smaller KV cache periodically during long-form generation.
    \item We evaluate our method comprehensively on various benchmarks and two LLMs, and conduct ablation studies on our design choices (e.g., dynamic stride vs. fixed stride cache updates). 
\end{itemize}



\section{RefreshKV for Long-Form Generation with Long-Context LLMs}
\label{methods}
\subsection{Background and Setting} \label{subsec:motivation}

Let $M$ be a language model and $\mathtt{x}$ be an input sequence of tokens, $x = x_1, \cdots x_L$. 
At inference time, $M$ generates an output token sequence $\hat{y}= y_1, \cdots y_N $ in two stages: (1) \textbf{Pre-filling stage} where $M$ ingests the input  and constructs the KV cache for all $L$ tokens, and (2) \textbf{Generation stage} where it samples one token $y_i$ at a time from the conditional distribution $P_M(y_i | x, y_1 \cdots y_{i-1})$. At each step, the model attends to tokens in the KV cache, and updates the cache to include the current token's key-value pairs.

Our goal is to reduce the inference latency during the generation stage without severe degradation of model performance. There are two main reasons for latency increase; first, the attention computation increases quadratically with input length $L$. Second, a large $L$ necessitates maintaining a large KV cache of the past tokens, incurring latency due to the full KV cache movement from the GPU HBM.\footnote{\citet{adnan2024keyformer} reports up to 40\% of the inference latency can be attributed to data movement.} 

Prior approaches, like H$_2$O \citep{zhang2024h2o} and SnapKV \citep{li2024snapkv}, address this by permanently evicting ``unimportant'' tokens during the decoding process to maintain a small KV cache. While such methods have shown to be effective for short-form generation task such as ``Needle-in-a-Haystack''(NIAH)~\citep{NIAH}, it has the potential downside of prematurely removing tokens useful for subsequent generation steps. Instead of this strict strategy, we propose to periodically \textit{update} the small KV cache by performing full attention over all the tokens in the context and constructing the small cache based on the attention pattern. As the cache is only occasionally updated, our method reduces both attention computation and data movement by attending to the small cache.

\begin{figure*}
    \centering
        \includegraphics[scale=0.30, trim=10mm 90mm 0mm 10mm]{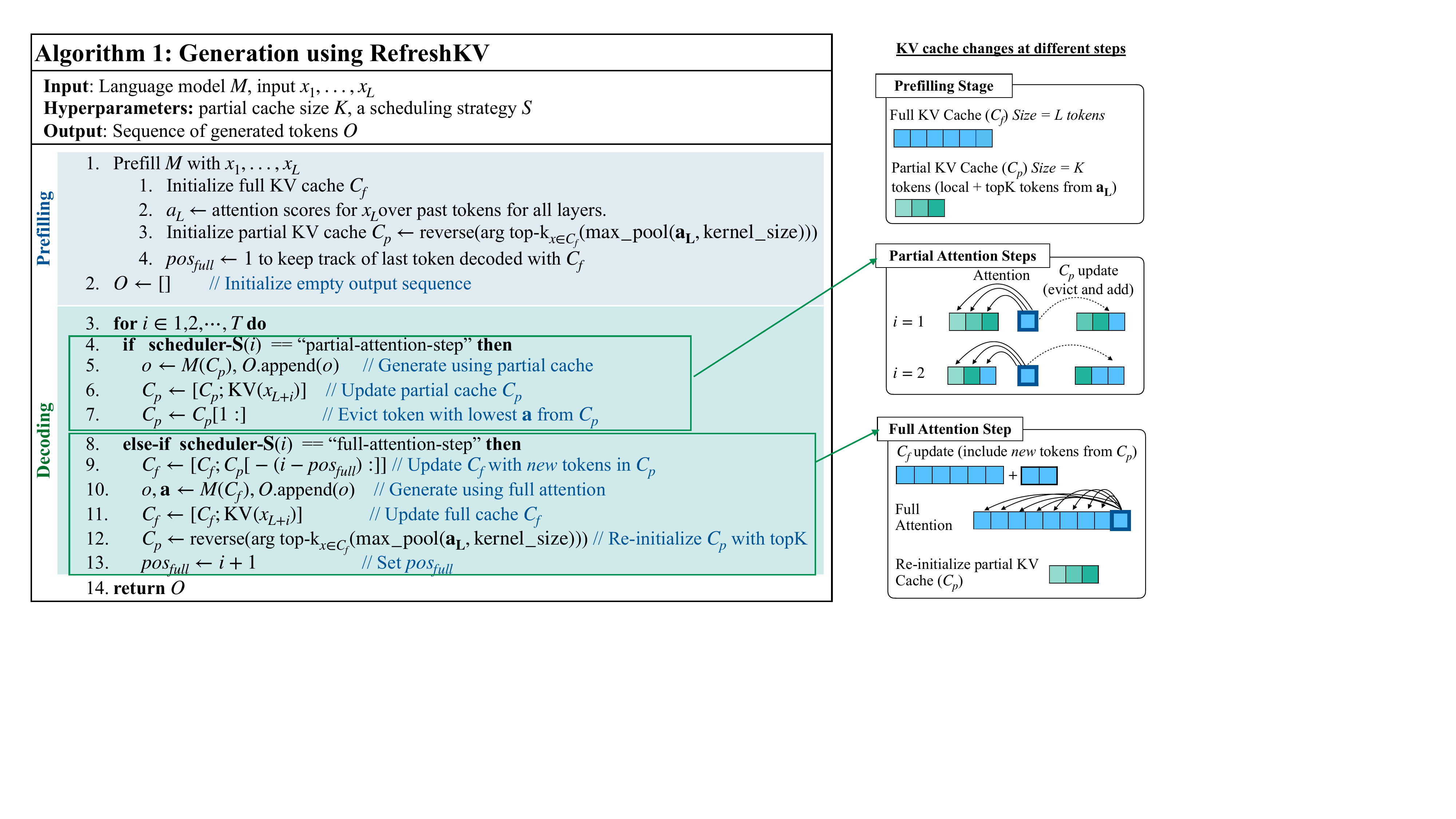}
    \vspace{-15pt}
    \caption{Pseudocode for \methodE. The model prefills the prompt with full attention and initialize the partial cache $C_{p}$ cache with attention scores of the last token. For each partial attention step, we decode with the partial cache and append the KV pairs of the input token to the partial cache. We evict the token with the lowest attention score to maintain a fixed-sized partial cache. For the full attention step, we first update the full KV cache with the new tokens decoded with the partial cache, then decode with the full cache and refresh the partial cache.} 
    \label{fig:algorithm}
\end{figure*}

\subsection{Methodology and Implementation}

We present the pseudocode for generating output tokens using \method in Figure~\ref{fig:algorithm}. The algorithm takes as input a language model $M$ and a sequence of input tokens $x_{1}, ... , x_{L}$. As a first step, we prefill $M$ with the input sequence. Then, we alternate between full and partial attention. Our approach maintains two separate KV caches $C_f$ and $C_p$, corresponding to KV cache used in the full and partial attention steps respectively. These three components of the algorithm are described below:

    \paragraph{Prefilling stage} (lines 1-2): Given input $x_{1}, ... , x_{L}$, we prefill with full attention $M$ and initialize full KV cache $C_f$ with $L$ tokens. We also obtain the attention scores $\mathbf{a_{L}}$ for the last token $x_L$. To determine the top K tokens to keep, we employ max pooling over attention scores of surrounding tokens, instead of the raw attention scores to preserve information completeness following prior work \citep{li2024snapkv}.\footnote{For models with Grouped Query Attention \citep{Ainslie2023GQATG}, we aggregate attention scores for all query heads in the same group by taking the max to identify the top K tokens. Our ablations (reported in  Table \ref{tab:query_aggregation_results} in the Appendix) show that taking the max outperforms other aggregation method such as mean, or relying solely on one of the query head in the group.} 

\paragraph{Deciding when to decode with full cache} (line 4): We need to decide when to alternate between performing attention over all tokens and performing attention over the smaller cache. 
One straightforward way is to use a fixed schedule, i.e. performing full attention every $S$ steps. However, this enforces the same schedule for \textit{all} the layers and input text. Instead, we propose an adaptive schedule based on the similarity between query vector of the current step and the query vector of the most recent full attention step. Intuitively, if the query vector of a particular layer and head for the current step is similar to the query vector of the most recent full attention step, the attention pattern should be similar. Thus, we only perform the full attention step when this similarity is lower than a threshold. 

Concretely, at every $S^{th}$ decode step, for each layer $l$,  we first determine whether we \emph{need} to perform full attention. We calculate the cosine similarity between the query vectors of the input token $t$ averaged across all query heads in layer $l$, with the averaged query vector of the most recent full attention step for that layer. If the similarity is higher than a threshold $s$, 
we decode with the partial cache $C_{p}$, and otherwise decode with $C_{f}$ for layer $l$. We describe details for each scenario below.
To minimize the computational overhead of the similarity check, we perform this only every $S$ steps; we call this query comparison (QC) stride.

    \paragraph{Decoding with partial cache} (lines 5-7):
    At each partial attention step, we generate the next token $y_t \sim M(C_p)$ using $C_{p}$ to compute attention and store the KV cache of the input token. This leads to a reduction in both the attention computation FLOPs and the latency due to KV cache movement (we only need to move the smaller KV cache $C_p$ instead of the larger full KV cache $C_f$, where $|C_p| << |C_f|$).  To maintain the size of $C_{p}$ as we decode each additional token and update the KV cache with this newly generated token, we remove the KV corresponding to the token with the lowest attention score in the full attention step from $C_p$ (line 7). We note that decoding with $C_{p}$ is equivalent to SnapKV~\citep{li2024snapkv} if the partial cache is never refreshed after prefilling.
    
    \paragraph{Decoding with full cache} (lines 9-13): At each full attention step, we first update the full KV cache $C_f$ with the key-value pairs of the tokens decoded with $C_{p}$. Next, we generate the next token $y_t \sim M(C_f)$ using the full KV cache $C_f$ and obtain the attention scores $\mathbf{a_{L}}$. Finally, we refresh the partial cache $C_p$ with the topK tokens based on $\mathbf{a_{L}}$.



\paragraph{Memory and Time requirements} Our method has memory requirement similar to that of vanilla attention, as we are not permanently evicting any tokens from the KV cache. 
However, our decoding latency is on par with other KV cache eviction methods, as later shown in our experiments in Section \ref{sec:exps}. We discuss memory and speed considerations in detail in Appendix~\ref{subsec:memtime}.


\section{Experiment Setup}
\label{sec:exps}

\paragraph{Models and Evaluation tasks} We evaluate our method with two long-context language models \llama~\citep{llama3} and \qwen~\citep{Yang2024Qwen2TR}. Both models can process inputs of up to 128K tokens. We conduct experiments on language modeling and downtream tasks:
\begin{itemize}[leftmargin=*]

    \item \textbf{Language modeling} We measure perplexity of the Arxiv and Book split of RedPajama \citep{together2023redpajama} with context size of 16K. We report results on 100 sequences for each domain. To simulate long-form generation, we report the perplexity of the last 256 tokens.
\item \textbf{Long-input, short output tasks:} We report the performance of {RULER}~\citep{hsieh2024ruler}, which consists of a set of 13 tasks with context size of 32K that require short output. 
\item \textbf{Long-input, long output tasks} We evaluate our methods on three sets of downstream tasks which require the model to generate long-form outputs (more than 100 tokens) given long-form inputs (more than 10k tokens).\footnote{For each dataset, we filter examples with input length <10K tokens. We report dataset statistics for each dataset in Table \ref{tab:dataset_statistics} in the Appendix.} (1) \textbf{long-context summarization tasks}: QMSum~\citep{zhong-etal-2021-qmsum}, GovReport~\citep{huang-etal-2021-efficient} and Novel Summarization~\citep{zhang-etal-2024-bench} and (2) \textbf{HTML to TSV} task from {LongProc}~\citep{ye25longproc} benchmark.\footnote{We exclude the other tasks from LongProc as they primarily involve short inputs, resulting in minimal speedup in our setting. For completeness, we report the performance of these tasks in Section \ref{sec:more_longproc} in the Appendix, observing a similar trend as the HTML to TSV task in terms of end-task performance.} We report results aggregated across three output lengths (0.5K, 2K, and 8K). We report ROUGE-L for summarization tasks and row-level F-1 score for the \textbf{HTML to TSV} task.


\item  \textbf{New task: Chain-of-key generation} We propose a synthetic task where model's previous generation steps, together with its long context input, guides future generation steps. Given a context which consists of a list of two-word keys, the model is tasked with generating a sequence of $T$ keys, such that the first word of the next key is the last word of the current key. This task requires models to look up information in the context based on what has been previously generated, resembling multi-hop retrieval. An example of the task is illustrated in Figure \ref{fig:RecycledAttention-Figure1}. We report accuracy of the output by the relative length of a valid chain (i.e. the length of the valid sub-chain divided by $T$). More details and examples are in Section \ref{subsec:chain_of_keys} in the Appendix.

\end{itemize}


\paragraph{Comparison systems} We implement the following baselines: (1)~\textit{\textbf{Vanilla}} attention that maintains and performs attention over the full KV cache (2)~\textbf{\textit{StreamingLLM}} \citep{Xiao2023EfficientSL} which consists  of ``sink tokens'' and recent tokens. 
(3)~\textbf{\textit{H$_2$O}} \citep{zhang2024h2o} which consists of recent tokens and dynamically updated ``heavy hitters'', defined by high cumulative attention scores. 
(4) \textbf{\textit{SnapKV}} \citep{li2024snapkv} which consists of tokens with high attention scores from the last few tokens in the prompt. We describe the setting for each baseline in Section \ref{sec:implementation_details} in the Appendix.

\paragraph{Inference settings} We prefill the model with the input and report wall clock times for the decoding phase. Our experiments are run on a single A100 80GB GPU using Flash Attention \citep{dao2023flashattention2}.\footnote{We describe implementation details in Section \ref{sec:implementation_details}.} We set $K$ to be 1/8 of the input length. NovelSumm contains the longest input length (100K tokens) and  we set $K$ to be 4096, corresponds to $1/25L$. We report results with greedy decoding. For \methodN, we report results for two different query comparison strides \{5, 10\} with a similarity threshold $s$ of 0.85 for \llama{} and 0.95 for \qwen{}. We determine the value of $s$ by experimenting with a range of values on a held-out set of the Book dataset (reported in Section \ref{sec:query_sim} in the Appendix) and apply the same threshold for all the tasks.

\section{Results}
\label{sec:results}

\begin{table}[t]
    %
\small
\setlength{\tabcolsep}{4.5pt}
\begin{center}
\begin{tabular}{@{}l|cc@{}}
\toprule

\textbf{Method} &  \textbf{Arxiv/Book PPL $\downarrow$} & \textbf{Time $\downarrow$} \\
\midrule

\multicolumn{3}{c}{\textit{Llama-3.1-8B}} \\
\rowcolor{LightGrey}Vanilla  & 2.22/7.07 & 7.50 \\
Streaming      & 2.62/7.94 & 6.61 \\
H$_2$O &   {2.48}/{7.60} & 10.77 \\
SnapKV &  2.54/7.78 & 6.77 \\
Refresh (QC=5) & \textbf{2.27}/\textbf{7.31} & 6.67 \\
Refresh (QC=10) & 2.32/7.41 & 6.33 \\
\midrule 

\multicolumn{3}{c}{\textit{QWEN-2-7B}} \\
\rowcolor{LightGrey}Vanilla  & 2.33/8.26 & 9.07 \\ 
Streaming      & 2.75/9.10 & 6.27 \\
H$_2$O& {2.68}/{9.02} & 11.57 \\
SnapKV & 2.80/9.18 & 6.09 \\ 
Refresh (QC=5) & \textbf{2.39}/\textbf{8.55} & 6.71 \\
Refresh (QC=10) & 2.49/8.72 & 6.33 \\
\bottomrule
\end{tabular} 
\end{center}
\caption{Perplexity results and latency on language modeling task for LLama-3.1-8B and QWEN-2-7B. We report results on Arxiv and Book corpora with input context length of $16K$ tokens. We set $K=2048$. }
\label{tab:lm_main_results}
\end{table}

\begin{table*}
\footnotesize
\setlength{\tabcolsep}{2.5pt}
\begin{center}
\begin{tabular}{@{}lc|c|c|c|c|c@{}}
\toprule
{\textbf{Input/Output length}} & 32K/$<$30 & 10K/ 0.1K & 10K/0.7K  & 128K/1K & 30K/2.2K & 22K/50 \\
\textbf{Dataset} & \textbf{RULER}  & \textbf{QMSum} & \textbf{GovReport} & \textbf{NovelSumm} & \textbf{HTML to TSV } & \textbf{Chain-of-key*} \\ 
\textbf{Method} &\textbf{Acc$\uparrow$} & \textbf{R-L$\uparrow$} & \textbf{R-L$\uparrow$} & \textbf{R-L$\uparrow$} & \textbf{F-1$\uparrow$} & \textbf{Acc$\uparrow$} \\ 
\midrule
\rowcolor{LightGrey}Vanilla & 90 / 79 &  25.63 / 24.98 & 34.11 / 33.38 & 31.29 / 19.91 & 33 / 24 & 56 / 83 \\ 
Streaming &  22 / 21 &  22.27 / 20.30 & 16.30 / 23.84  & 24.66 / \textbf{22.11} & 2  / 5 & 2 / 2\\ 
H$_2$O &   21 / 21 &  {22.12} / 20.83 & 27.41 / 26.91 & 19.31 / 18.51 & 0 / 0 & 10 / 11 \\
SnapKV  & 79 / 58 & 24.33 / 22.93 & 28.06 / 28.80 & 29.23 / 19.09 & 0 / 0 & 12 / 13  \\ 
RefreshKV (QC=5) &  \textbf{86} / \textbf{75} & \textbf{24.92} / \textbf{24.34} &  \textbf{32.56} / \textbf{31.40}  &  \textbf{29.98} / 19.70 &  \textbf{17} / \textbf{10} & \textbf{25 / 24} \\
RefreshKV (QC=10) & 80 / 67 &   24.73  / 23.98 & 31.47 / 31.36  & 29.37 / 18.94  &  8 / 6   & 15 / 15 \\
\bottomrule
\end{tabular} 
\end{center}
\vspace{-1em}
\caption{Downstream task performance. In each cell, the first number represents the performance of Llama-3.1 model and the second number for QWEN-2 model. *We report performance of Llama-3.1-70B and Qwen-2-72B for the chain-of-key task, as the smaller variants cannot perform the task even in vanilla setting.} 
\label{tab:end_tasks_results}
\end{table*}


\subsection{Language Modeling}
\label{sec:lm}

\begin{figure}
  \begin{center}
    \includegraphics[scale=0.16,trim=0 0mm 0mm 0mm]{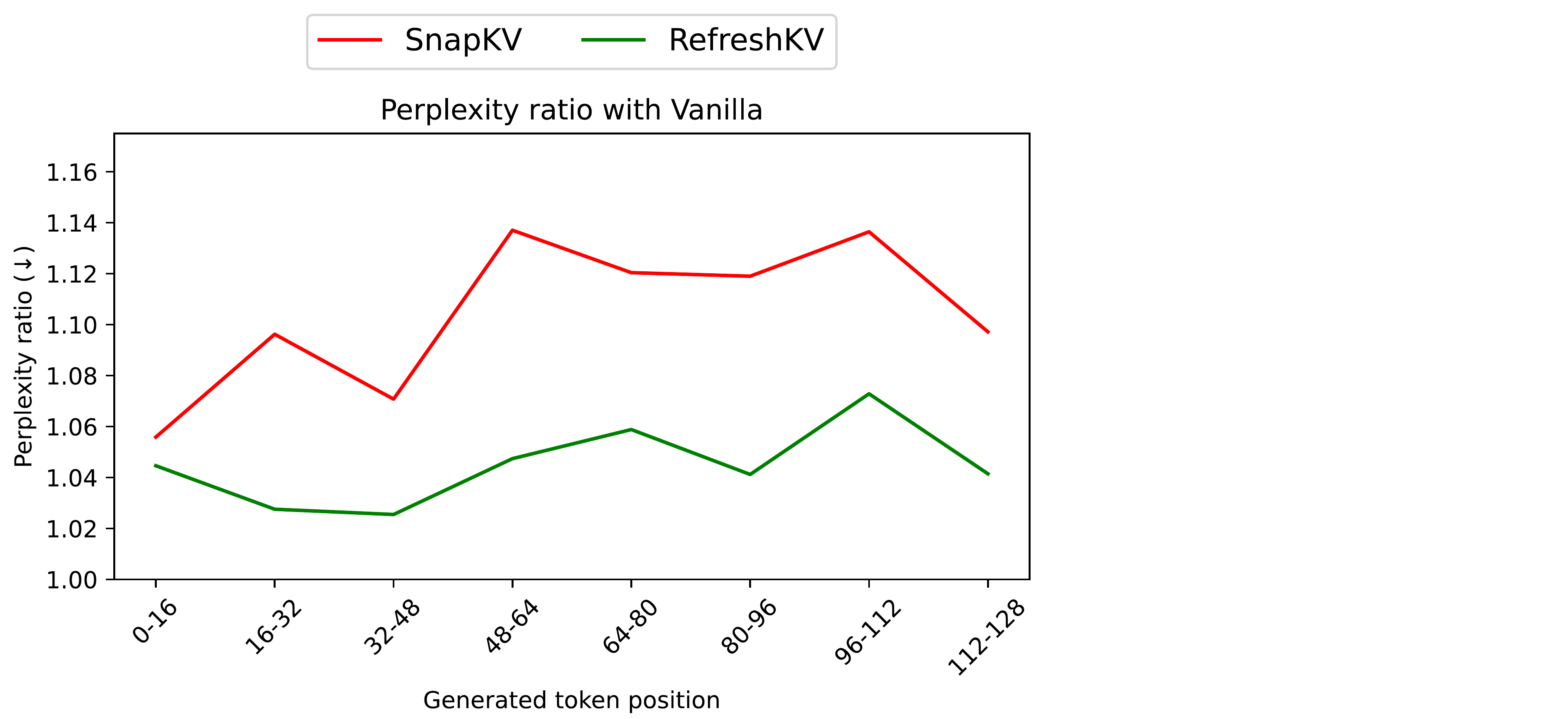}
  \vspace{-18pt}
  \end{center}
  \caption{We plot the perplexity ratio against the vanilla baseline for RefreshKV (with stride of 10) and SnapKV based on the tokens generated (x axis). While the ratio is similar at the beginning of the sequence, as the generation goes SnapKV's perplexity diverges from vanilla approach while that of RefreshKV is relatively stable.
  }\vspace{-1em}
\label{fig:perplexity_ratio}
\end{figure}

Table~\ref{tab:lm_main_results} outlines the performance of the baselines and \methodN~for perplexity. For both models, \methodN~achieves better perplexity and comparable inference speeds compared to StreamingLLM and SnapKV  for $QC=10$. Our method also achieves better performance than the best baseline, H$_2$O,  with a much shorter inference time per example, as we do not require accessing attention score at each decoding step. Setting $QC=5$ increases inference time but also brings performance gain compared to $QC=10$, allowing a performance-efficiency trade-off.


The key distinction between \methodN~and SnapKV is that our method \textit{refreshes} the partial cache as generation progresses. We compare the perplexity degradation ratio of both methods relative to vanilla attention over different generation timestamps in Figure \ref{fig:perplexity_ratio} with \llama{} on the book dataset. While both methods begin with a similar perplexity ratio compared to vanilla (step 0-16), SnapKV's performance degrades as generation proceeds, whereas \methodN~maintains a stable ratio, highlighting the benefit of refreshing the small KV cache during generation.

\subsection{Downstream Tasks}
\label{sec:ruler}
Results for downstream tasks are reported in Table \ref{tab:end_tasks_results}. We also report the average input and output length for each dataset. For {RULER}, we report results aggregated over 13 tasks here and report the per-task performance in Table \ref{tab:ruler_detail} in the Appendix. 

\paragraph{Eviction-based methods fail for long-form generation tasks.} Baseline methods that evict tokens from the KV cache permanently (StreamingLLM, H$_2$O and SnapKV) show degradation for tasks that require long-form outputs. While SnapKV performs better than the other two baselines on {RULER}, it shows severe performance degradation on the {HTML to TSV} task, achieving 0 F-1 scores for the former. For the {Chain of key} task, eviction-based methods are  unable to generate a chain with more than two keys, achieving accuracy < 20.

\paragraph{\method closes the gap between vanilla and eviction-based approach.} On {HTML to TSV} task, \methodN~with $QC=5$ recovers 52\% and 42\% of performance for \llama{} and \qwen{} respectively. On the \textbf{Chain-of-key} task, \methodN~is the only method that is able to generate a valid key with length longer than two keys, as shown in Figure \ref{fig:RecycledAttention-Figure1}. For the long-form summarization tasks, \methodN~outperforms baselines in all three datasets, except for NovelSumm with \qwen{}, where StreamingLLM outperforms the vanilla full attention. We also observe gains for RULER tasks, particularly the subtasks that require generating longer output (e.g. generating multiple keys), which we discuss in Section \ref{sec:ruler_finegrained} in the Appendix.






\section{Ablation Studies}\label{sec:ablation}

\begin{table*}
\begin{center}
\footnotesize
\begin{tabular}{@{}l|ccc|ccc|ccc@{}}
\toprule
 \textbf{Schedule}  & \multicolumn{3}{c|}{\textbf{Book}} & \multicolumn{3}{c|}{\textbf{HTML (0.5K)}}  & \multicolumn{3}{c}{\textbf{GovReport}} \\  \midrule
& \textbf{Time $\downarrow$} &  \textbf{Stride}  & \textbf{PPL $\downarrow$} & \textbf{Time $\downarrow$} & \textbf{Stride} & \textbf{Acc $\uparrow$}  & \textbf{Time $\downarrow$} &  \textbf{Stride} & \textbf{R-L $\uparrow$}  \\ \midrule
\multicolumn{9}{c}{\textit{Llama-3.1-8B}} \\ 
 \rowcolor{LightGrey} Vanilla & 7.50 & - & 7.07 & 1.52 & - & 43 & 1.43 & - & 34.11 \\
Fixed  & 7.20 &  10 & 7.40 & 1.40 & 10 & 17 & 1.37 & 10 & {32.30} \\
Dynamic (QC=5, s=0.85) & 7.17 &  12 & \textbf{7.31} & 1.40 & 14 & \textbf{30}  & 1.38 & 14 & \textbf{32.56} \\
Fixed   & 6.99 & 15 & 7.45 &  1.37 & 15 &  8 & \textbf{1.34} &  15 & 30.67     \\
Dynamic (QC=10, s=0.85)  & \textbf{6.89} & 17 & 7.41 & \textbf{1.33} &  19 & 16 & \textbf{1.34} &  19 & 31.47 \\ \midrule
\multicolumn{9}{c}{\textit{Qwen-2-7B}} \\
 \rowcolor{LightGrey} Vanilla & 9.07 & - & 8.26 & 1.96 & - & 35 & 1.73 & - & 33.38 \\
Fixed  & 6.59 & 10 & 8.74 & 1.38 &  10 &  8 &  1.29 & 10 & {31.18}  \\ 
Dynamic (QC=5, s=0.95) & 6.71 & 7 & \textbf{8.55} &  1.29 & 7 & \textbf{20}  &  1.34 &  7 & \textbf{31.40}  \\ 
Fixed  & {6.43} & 15 & 8.81 & 1.31 & 15 & {9} & \textbf{1.27} &  15 & 30.73 \\
Dynamic (QC=10, s=0.95) & \textbf{6.33} &  11 &  8.72 &  \textbf{1.23} &  12 & 14  & 1.28 & 12 & 31.36  \\ 
\bottomrule
\end{tabular} 
\end{center}
\vspace{-0.7em}
\caption{Results comparing fixed stride and dynamic stride based on query similarity. In all tasks, dynamic stride shows better task performance while performing full attention step fewer times.}
\label{tab:dynamic_stride}
\end{table*}



\begin{table}
\begin{center}
\footnotesize
\begin{tabular}{@{}llcc@{}}
\toprule
\textbf{Method} & \textbf{Stride} & \textbf{Arxiv} & \textbf{HTML (0.5K)}   \\ 
\midrule
\rowcolor{LightGrey} Vanilla & -  & 2.22 & 43 \\  
SnapKV & - & 2.54 & 0 \\ 
 RefreshKV & 10 & 2.32 & 16 \\ 
  - w/o refresh & 10 & 2.50 & 0  \\ 
 - w/o full attention & 10 & 2.32 & 16 \\ 

\bottomrule
\end{tabular} 
\end{center}
\caption{Ablation study on LLama-3.1-8B. We report perplexity for Arxiv and F-1 score for the HTML to TSV task.}
\label{tab:ablation_study}\vspace{-0.5em}
\end{table}

\subsection{Adaptive stride vs. Fixed stride}
We trigger full attention step when the query vector of the input token is substantially different from the query vector of the most recent full attention step. Can we use a simpler strategy to decide when to perform full attention? In this section, we explore refreshing at a fixed stride, performing full attention every $N$-th step across all the layers. 

\paragraph{Setting} We compare the results of (1) employing a dynamic stride with the set-up in Section \ref{sec:exps}, i.e. $QC$ stride of \{5, 10\} and similarity threshold $s=0.85$ (\llama{}) and $s=0.95$ (\qwen{}) and (2) employing a fixed stride $S$ of \{10, 15\} for comparable decoding time. We report results on the language modeling task on the {Book} dataset and two downstream tasks. We report the decoding time measured on one A100 machine. For the language modeling task, we report the time for generating 256 tokens. For the downstream tasks, we measure the time of generating the first 50 tokens. We also report the \textit{effective} stride averaged across all the layers, i.e. how often is full attention performed when employing dynamic strides.

\paragraph{Results} Table~\ref{tab:dynamic_stride} presents the results.  For \llama{}, comparing $QC=5$ and $S=10$, employing dynamic stride consistently achieves better performance with similar or less decoding time for all three tasks. We see a similar trend comparing $QC=10$ and $S=15$. For \qwen{}, dynamic stride achieves better performance across all three tasks, with slightly more decoding time on Govreport. We also observe slightly different effective stride for different tasks when employing the same $QC$ and $s$, showing that dynamic stride enable flexible scheduling based on the context. We report per-layer stride in Section \ref{subsec:effective_stride} in the Appendix.

\subsection{Impact of full attention steps} Compared to other baseline methods which never perform full attention during the generation, \methodN~involves extra attention calculation (i.e. attending over the entire output). To tease apart the performance gains from occasional full attention step and updating the small KV, we present two ablation setting for \methodN: (1) \textit{w/o refresh} which performs attention over the full KV cache at the fixed stride of $S$ but without refreshing the partial cache. This is equivalently using the partial cache obtained with SnapKV and occasionally performing full attention. (2) \textit{w/o full attention} which calculates the attention scores over the entire KV cache and updates the partial cache, then attends to the updated partial cache, instead of attending to the full KV cache, at stride $S$.

Results are in Table \ref{tab:ablation_study}. While performing occasional full attention (\textbf{w/o refresh}) improve perplexity slightly compared to SnapKV, the performance lags behind \methodN. In contrast, the ablation setting where partial cache is refreshed (\textbf{w/o full attention}) achieves the same performance of \methodN~ for both tasks. This shows that the gain of \methodN~mostly comes from refreshing the partial KV cache, instead of performing occasion full attention over the entire cache.


\section{Continued Pre-training with RefreshKV}
\label{subsec:cpt}

We have demonstrated \method can be used as an inference-time method. However, since the LLMs we study are trained with full attention, applying \methodN~during inference introduces a discrepancy between training and inference. Specifically, it involves attending to a non-contiguous sequence of tokens in the partial cache. Here, we explore continued pretraining with RefreshKV to adapt models to this new attention pattern.

To make training setting simpler, we do not fully implement RefreshKV during training. We use a fixed stride of 50 and never refresh the partial cache. We assume a length $L+S$ for all sequences, where $L$ is the pre-fill length. We perform standard attention over all past tokens for the first $L$ tokens. We emulate the partial attention pattern for the last $S$ tokens in the sequence during training.  For the next $S$ tokens, we perform attention over the top K tokens identified as well as local tokens (i.e. tokens $L+1$ onwards).  We train the model with next token prediction loss for all the tokens in the sequence.  

\paragraph{Setup} We set $L=8092$, $S=50$ and $K=2048$ for this experiment. We randomly sample a subset of 200k sequences from the Arxiv split of RedPajama dataset. We split the data into 80\%, 10\% and 10\% train/dev/test splits, resulting in 120k training data samples.  We perform continued pre-training on \llama{} and describe implementation details in Section \ref{sec:implementation_details} in the Appendix.

\begin{table}
\begin{center}
\footnotesize
\begin{tabular}{@{}lrcc@{}}
\toprule
\textbf{Method} & \textbf{Stride} & \textbf{Test PPL (8K)} & \textbf{Test PPL (16K)}   \\ 
\midrule
\rowcolor{LightGrey} Vanilla & -  & 2.70 $\rightarrow$ 2.70 & 2.50 $\rightarrow$ 2.50  \\ 
 Streaming& - &3.40 $\rightarrow$ 3.38 &  3.50 $\rightarrow$  3.49 \\ 
 H$_2$O & - & 3.95 $\rightarrow$ 3.90 & 3.52 $\rightarrow$ 3.49 \\ 
 SnapKV & -  & 3.21 $\rightarrow$ 3.15 & 2.98 $\rightarrow$  2.92 \\ 
RefreshKV & 10  &  2.83 $\rightarrow$ 2.79 & 2.57 $\rightarrow$ 2.56 \\ 
RefreshKV & 25  &  2.97 $\rightarrow$ 2.93 & 2.67 $\rightarrow$ 2.63 \\ 
 RefreshKV & 50 &  3.13 $\rightarrow$ 3.05 &  2.79 $\rightarrow$ 2.72 \\ 
\bottomrule
\end{tabular} 
\end{center}
\caption{Results on continued pre-training with RefreshKV for LLaMA-3.1. The context size is 8k and we report perplexity on the last 50 tokens. We report the performance for each setting before (the number on the left) and after (the number on the right) CPT.}
\label{tab:cpt_results}\vspace{-0.5em}
\end{table}

\paragraph{Evaluation} As our continued pretraining is relatively small scale on the base model, we focus on evaluating on the language modeling task for two settings: (1) input size $L=$8K consistent with the training set-up and (2) $L=$16K. We set $K=1/8L$ For each method, we report the performance from the pre-trained checkpoint  and the performance after continued pre-training.  

\paragraph{Results} We report the results in Table \ref{tab:cpt_results}, each row represents a different inference strategy on the same model. Despite the mismatch in how partial KV was constructed, continued pre-training benefits other methods (Streaming, H$_2$O) slightly. We see larger gain for \methodN~from continued pre-training across all settings. Our training assumes a fixed stride of 50, but we see performance gain for different strides ($S=10, 25$). Training on shorter context (8K) also translates to gains when inferencing on longer context (16K), showing promise for improving the performance of \methodN~with continued pre-training.

\section{Related Work} \label{related_works}

\paragraph{Efficient inference methods}  
Various techniques have been proposed to enhance inference efficiency, which are orthogonal to and can be combined with our approach. 
FlashAttention~\citep{dao2023flashattention2} achieves significant gain in inference speed by optimizing attention computations on GPUs. A line of work~\citep{Xiao2022SmoothQuantAA, Liu2024KIVIAT, hooper2024kvquant} proposes to quantize KV caches to reduce both memory and computation cost. 

\paragraph{KV cache eviction} Recent work extensively studies KV cache eviction strategies, such as keeping only ``sink'' and recent tokens in the KV cache \citep{Xiao2023EfficientSL}; or tokens with high accumulative attention scores \citep{zhang2024h2o}.  A line of work propose query-aware eviction strategies, using the attention scores of the last few tokens in the prompt to select tokens to keep \citep{li2024snapkv, chen-etal-2024-nacl}. Other works design eviction strategies based on attention patterns of different heads \citep{ge2024model, xiao2024duo} or different layers \citep{cai2024pyramidkv, yang-etal-2024-pyramidinfer}. We show that such eviction-based methods can fail on long-form generation tasks and propose to refresh the small KV cache during generation.


\paragraph{Sparse attention} Our method achieves efficiency by performing sparse attention. Earlier work \citep{Zaheer2020BigBT, Beltagy2020Longformer} investigates training LLMs with a fixed sparse attention pattern (such as a sliding window) to reduce computational complexity. Training-free methods such as Unlimiformer \citep{bertsch2023unlimiformer} and InfLLM \citep{xiao2024infllm} performs attentions on subset of tokens which received the highest attention scores, with the goal of extending the context window of a given language model. In contrast, we leverage previous tokens' attention scores to select tokens to attend to for long-context models, which can already handle sequences with up to 128k tokens. MInference \citep{jiang2024minference} identify head-specific patterns to perform sparse attention, focusing on accelerating the prefilling stage.  Similar to ours, SparQ \citep{pmlr-v235-ribar24a} and  Quest~\citep{tang2024quest} achieves decoding time speed-up by attending to subset of tokens. Instead of leveraging the attention patterns of previous tokens, these methods build specialized kernel to approximate attention and identify critical tokens.



\section{Conclusion}
We propose \methodE, an inference-time method which accelerate long-form generation for long-context input by decoding from a small, dynamic KV cache that is updated based on attention patterns of neighboring tokens.
Compared to previous work which permanently evict tokens from the context, \methodE~ maintains the full KV cache and alternates between inferencing over the full and small KV cache. We apply our method to two off-the-shelf long-context model and show that our method reduces inference wall-clock time while better preserving performance compared to eviction-based methods on long-form generation tasks. Finally, we show that continued pre-training the model with \method can further improve the performance-efficiency trade-off.

\section*{Limitations}

\paragraph{Proposed method}
While we focus on accelerating inference speed, our method does not reduce memory requirement for using long-context LLMs, which can be a bottleneck for certain use cases. Our objective is to accelerate decoding for long-context models. While our method outperforms eviction-based approaches, it still involves a trade-off between performance and efficiency. In this study, we employ query similarity based dynamic schduling to decide when to perform full attention and refresh the small KV cache. Future work can explore other strategy, such as more exhausively tuning the similarity threshold, or setting a different threshold per layer.

\paragraph{Experimental settings}
We have conducted experiment with two open-sourced long-context models and two evaluation tasks setting. We did not test out more language models and other long-context benchmarks~\citep{an2023levalinstitutingstandardizedevaluation,karpinska2024thousandpairsnovelchallenge} given our limited compute resources. For the same reason, our experiment on continued pre-training is relatively small scale on a limited domain. We have demonstrated the effectiveness of refreshing a small KV cache constructed with attention scores and use the same size across different layers. Future work can extend our method to refresh smaller cache constructed with different strategy, e.g. layer-specific strategies~\citep{yang-etal-2024-pyramidinfer, cai2024pyramidkv}. Finally, our method is not limited to the language domain. Future work can explore applying \methodN~to other modalities, for example, vision transformers.

\section*{Acknowledgments}

We thank Xi Ye, Wenting Zhao and the UT NLP group for helpful feedback. We also thank reviewers for helpful feedback for an earlier manuscript. The work is partially supported by a gift from Apple. This work was done in part while the first and last author was visiting the Simons Institute for the Theory of Computing. 

\bibliography{custom}

\begin{thebibliography}{38}
\providecommand{\natexlab}[1]{#1}

\bibitem[{Adnan et~al.(2024)Adnan, Arunkumar, Jain, Nair, Soloveychik, and Kamath}]{adnan2024keyformer}
Muhammad Adnan, Akhil Arunkumar, Gaurav Jain, Prashant Nair, Ilya Soloveychik, and Purushotham Kamath. 2024.
\newblock Keyformer: Kv cache reduction through key tokens selection for efficient generative inference.
\newblock \emph{Proceedings of Machine Learning and Systems}, 6:114--127.

\bibitem[{Ainslie et~al.(2023)Ainslie, Lee-Thorp, de~Jong, Zemlyanskiy, Lebr'on, and Sanghai}]{Ainslie2023GQATG}
Joshua Ainslie, James Lee-Thorp, Michiel de~Jong, Yury Zemlyanskiy, Federico Lebr'on, and Sumit~K. Sanghai. 2023.
\newblock \href {https://api.semanticscholar.org/CorpusID:258833177} {Gqa: Training generalized multi-query transformer models from multi-head checkpoints}.
\newblock \emph{ArXiv}, abs/2305.13245.

\bibitem[{An et~al.(2023)An, Gong, Zhong, Zhao, Li, Zhang, Kong, and Qiu}]{an2023levalinstitutingstandardizedevaluation}
Chenxin An, Shansan Gong, Ming Zhong, Xingjian Zhao, Mukai Li, Jun Zhang, Lingpeng Kong, and Xipeng Qiu. 2023.
\newblock \href {https://arxiv.org/abs/2307.11088} {L-eval: Instituting standardized evaluation for long context language models}.
\newblock \emph{Preprint}, arXiv:2307.11088.

\bibitem[{Bai et~al.(2023)Bai, Lv, Zhang, Lyu, Tang, Huang, Du, Liu, Zeng, Hou, Dong, Tang, and Li}]{bai2023longbench}
Yushi Bai, Xin Lv, Jiajie Zhang, Hongchang Lyu, Jiankai Tang, Zhidian Huang, Zhengxiao Du, Xiao Liu, Aohan Zeng, Lei Hou, Yuxiao Dong, Jie Tang, and Juanzi Li. 2023.
\newblock Longbench: A bilingual, multitask benchmark for long context understanding.
\newblock \emph{arXiv preprint arXiv:2308.14508}.

\bibitem[{Beltagy et~al.(2020)Beltagy, Peters, and Cohan}]{Beltagy2020Longformer}
Iz~Beltagy, Matthew~E. Peters, and Arman Cohan. 2020.
\newblock Longformer: The long-document transformer.
\newblock \emph{arXiv:2004.05150}.

\bibitem[{Bertsch et~al.(2023)Bertsch, Alon, Neubig, and Gormley}]{bertsch2023unlimiformer}
Amanda Bertsch, Uri Alon, Graham Neubig, and Matthew Gormley. 2023.
\newblock \href {https://arxiv.org/abs/2305.01625} {Unlimiformer: Long-range transformers with unlimited length input}.
\newblock In \emph{Advances in Neural Information Processing Systems}, volume~36, pages 35522--35543. Curran Associates, Inc.

\bibitem[{Cai et~al.(2024)Cai, Zhang, Gao, Liu, Liu, Lu, Xiong, Dong, Chang, Hu et~al.}]{cai2024pyramidkv}
Zefan Cai, Yichi Zhang, Bofei Gao, Yuliang Liu, Tianyu Liu, Keming Lu, Wayne Xiong, Yue Dong, Baobao Chang, Junjie Hu, et~al. 2024.
\newblock Pyramidkv: Dynamic kv cache compression based on pyramidal information funneling.
\newblock \emph{CoRR}.

\bibitem[{Chen et~al.(2024)Chen, Wang, Shang, Cui, Zhang, Liu, Wang, Sun, Yu, and Wu}]{chen-etal-2024-nacl}
Yilong Chen, Guoxia Wang, Junyuan Shang, Shiyao Cui, Zhenyu Zhang, Tingwen Liu, Shuohuan Wang, Yu~Sun, Dianhai Yu, and Hua Wu. 2024.
\newblock \href {https://doi.org/10.18653/v1/2024.acl-long.428} {{NACL}: A general and effective {KV} cache eviction framework for {LLM} at inference time}.
\newblock In \emph{Proceedings of the 62nd Annual Meeting of the Association for Computational Linguistics (Volume 1: Long Papers)}, pages 7913--7926, Bangkok, Thailand. Association for Computational Linguistics.

\bibitem[{Child et~al.(2019)Child, Gray, Radford, and Sutskever}]{Child2019GeneratingLS}
Rewon Child, Scott Gray, Alec Radford, and Ilya Sutskever. 2019.
\newblock \href {https://api.semanticscholar.org/CorpusID:129945531} {Generating long sequences with sparse transformers}.
\newblock \emph{ArXiv}, abs/1904.10509.

\bibitem[{Dao(2024)}]{dao2023flashattention2}
Tri Dao. 2024.
\newblock Flash{A}ttention-2: Faster attention with better parallelism and work partitioning.
\newblock In \emph{International Conference on Learning Representations (ICLR)}.

\bibitem[{Dettmers et~al.(2021)Dettmers, Lewis, Shleifer, and Zettlemoyer}]{bitsandbytes}
Tim Dettmers, Mike Lewis, Sam Shleifer, and Luke Zettlemoyer. 2021.
\newblock \href {https://arxiv.org/abs/2110.02861} {8-bit optimizers via block-wise quantization}.
\newblock \emph{CoRR}, abs/2110.02861.

\bibitem[{Ge et~al.(2024)Ge, Zhang, Liu, Zhang, Han, and Gao}]{ge2024model}
Suyu Ge, Yunan Zhang, Liyuan Liu, Minjia Zhang, Jiawei Han, and Jianfeng Gao. 2024.
\newblock \href {https://openreview.net/forum?id=uNrFpDPMyo} {Model tells you what to discard: Adaptive {KV} cache compression for {LLM}s}.
\newblock In \emph{The Twelfth International Conference on Learning Representations}.

\bibitem[{Gemini(2024)}]{team2024google}
Gemini. 2024.
\newblock Google. gemini 1.5: Unlocking multimodal understanding across millions of tokens of context.
\newblock \emph{arXiv preprint arXiv:2403.05530}.

\bibitem[{Hooper et~al.(2024)Hooper, Kim, Mohammadzadeh, Mahoney, Shao, Keutzer, and Gholami}]{hooper2024kvquant}
Coleman Hooper, Sehoon Kim, Hiva Mohammadzadeh, Michael~W Mahoney, Yakun~Sophia Shao, Kurt Keutzer, and Amir Gholami. 2024.
\newblock Kvquant: Towards 10 million context length llm inference with kv cache quantization.
\newblock \emph{arXiv preprint arXiv:2401.18079}.

\bibitem[{Hsieh et~al.(2024)Hsieh, Sun, Kriman, Acharya, Rekesh, Jia, Zhang, and Ginsburg}]{hsieh2024ruler}
Cheng-Ping Hsieh, Simeng Sun, Samuel Kriman, Shantanu Acharya, Dima Rekesh, Fei Jia, Yang Zhang, and Boris Ginsburg. 2024.
\newblock Ruler: What's the real context size of your long-context language models?
\newblock \emph{arXiv preprint arXiv:2404.06654}.

\bibitem[{Huang et~al.(2021)Huang, Cao, Parulian, Ji, and Wang}]{huang-etal-2021-efficient}
Luyang Huang, Shuyang Cao, Nikolaus Parulian, Heng Ji, and Lu~Wang. 2021.
\newblock \href {https://doi.org/10.18653/v1/2021.naacl-main.112} {Efficient attentions for long document summarization}.
\newblock In \emph{Proceedings of the 2021 Conference of the North American Chapter of the Association for Computational Linguistics: Human Language Technologies}, pages 1419--1436, Online. Association for Computational Linguistics.

\bibitem[{Jiang et~al.(2024)Jiang, Li, Zhang, Wu, Luo, Ahn, Han, Abdi, Li, Lin, Yang, and Qiu}]{jiang2024minference}
Huiqiang Jiang, Yucheng Li, Chengruidong Zhang, Qianhui Wu, Xufang Luo, Surin Ahn, Zhenhua Han, Amir~H Abdi, Dongsheng Li, Chin-Yew Lin, Yuqing Yang, and Lili Qiu. 2024.
\newblock Minference 1.0: Accelerating pre-filling for long-context llms via dynamic sparse attention.
\newblock \emph{arXiv preprint arXiv:2407.02490}.

\bibitem[{Kamradt(2023)}]{NIAH}
Gregory Kamradt. 2023.
\newblock \href {https: //github.com/gkamradt/LLMTest NeedleInAHaystack/tree/main.} {Needle in a haystack - pressure testing llms, commercially usable llms}.

\bibitem[{Karpinska et~al.(2024)Karpinska, Thai, Lo, Goyal, and Iyyer}]{karpinska2024thousandpairsnovelchallenge}
Marzena Karpinska, Katherine Thai, Kyle Lo, Tanya Goyal, and Mohit Iyyer. 2024.
\newblock \href {https://arxiv.org/abs/2406.16264} {One thousand and one pairs: A "novel" challenge for long-context language models}.
\newblock \emph{Preprint}, arXiv:2406.16264.

\bibitem[{Li et~al.(2024)Li, Huang, Yang, Venkitesh, Locatelli, Ye, Cai, Lewis, and Chen}]{li2024snapkv}
Yuhong Li, Yingbing Huang, Bowen Yang, Bharat Venkitesh, Acyr Locatelli, Hanchen Ye, Tianle Cai, Patrick Lewis, and Deming Chen. 2024.
\newblock \href {https://openreview.net/forum?id=poE54GOq2l} {Snap{KV}: {LLM} knows what you are looking for before generation}.
\newblock In \emph{The Thirty-eighth Annual Conference on Neural Information Processing Systems}.

\bibitem[{Liu et~al.(2024)Liu, Yuan, Jin, Zhong, Xu, Braverman, Chen, and Hu}]{Liu2024KIVIAT}
Zirui Liu, Jiayi Yuan, Hongye Jin, Shaochen Zhong, Zhaozhuo Xu, Vladimir Braverman, Beidi Chen, and Xia Hu. 2024.
\newblock \href {https://api.semanticscholar.org/CorpusID:267413049} {Kivi: A tuning-free asymmetric 2bit quantization for kv cache}.
\newblock \emph{ArXiv}, abs/2402.02750.

\bibitem[{Meta(2024)}]{llama3}
Meta. 2024.
\newblock \href {https://api.semanticscholar.org/CorpusID:271571434} {The llama 3 herd of models}.
\newblock \emph{ArXiv}, abs/2407.21783.

\bibitem[{MosaicML(2023)}]{MosaicML2023Introducing}
NLP~Team MosaicML. 2023.
\newblock \href {https://www.mosaicml.com/blog/mpt-7b} {Introducing mpt-7b: A new standard for open-source, commercially usable llms}.

\bibitem[{Ribar et~al.(2024)Ribar, Chelombiev, Hudlass-Galley, Blake, Luschi, and Orr}]{pmlr-v235-ribar24a}
Luka Ribar, Ivan Chelombiev, Luke Hudlass-Galley, Charlie Blake, Carlo Luschi, and Douglas Orr. 2024.
\newblock \href {https://proceedings.mlr.press/v235/ribar24a.html} {{S}par{Q} attention: Bandwidth-efficient {LLM} inference}.
\newblock In \emph{Proceedings of the 41st International Conference on Machine Learning}, volume 235 of \emph{Proceedings of Machine Learning Research}, pages 42558--42583. PMLR.

\bibitem[{Tang et~al.(2024)Tang, Zhao, Zhu, Xiao, Kasikci, and Han}]{tang2024quest}
Jiaming Tang, Yilong Zhao, Kan Zhu, Guangxuan Xiao, Baris Kasikci, and Song Han. 2024.
\newblock \href {https://arxiv.org/abs/2406.10774} {Quest: Query-aware sparsity for efficient long-context llm inference}.
\newblock \emph{Preprint}, arXiv:2406.10774.

\bibitem[{Together(2023)}]{together2023redpajama}
Together. 2023.
\newblock \href {https://github.com/togethercomputer/RedPajama-Data} {Redpajama: an open dataset for training large language models}.

\bibitem[{Xiao et~al.(2024{\natexlab{a}})Xiao, Zhang, Han, Xiao, Lin, Zhang, Liu, Han, and Sun}]{xiao2024infllm}
Chaojun Xiao, Pengle Zhang, Xu~Han, Guangxuan Xiao, Yankai Lin, Zhengyan Zhang, Zhiyuan Liu, Song Han, and Maosong Sun. 2024{\natexlab{a}}.
\newblock Infllm: Unveiling the intrinsic capacity of llms for understanding extremely long sequences with training-free memory.
\newblock \emph{arXiv}.

\bibitem[{Xiao et~al.(2022)Xiao, Lin, Seznec, Demouth, and Han}]{Xiao2022SmoothQuantAA}
Guangxuan Xiao, Ji~Lin, Mickael Seznec, Julien Demouth, and Song Han. 2022.
\newblock \href {https://api.semanticscholar.org/CorpusID:253708271} {Smoothquant: Accurate and efficient post-training quantization for large language models}.
\newblock \emph{ArXiv}, abs/2211.10438.

\bibitem[{Xiao et~al.(2024{\natexlab{b}})Xiao, Tang, Zuo, Guo, Yang, Tang, Fu, and Han}]{xiao2024duo}
Guangxuan Xiao, Jiaming Tang, Jingwei Zuo, Junxian Guo, Shang Yang, Haotian Tang, Yao Fu, and Song Han. 2024{\natexlab{b}}.
\newblock Duoattention: Efficient long-context llm inference with retrieval and streaming heads.
\newblock \emph{arXiv}.

\bibitem[{Xiao et~al.(2023)Xiao, Tian, Chen, Han, and Lewis}]{Xiao2023EfficientSL}
Guangxuan Xiao, Yuandong Tian, Beidi Chen, Song Han, and Mike Lewis. 2023.
\newblock \href {https://api.semanticscholar.org/CorpusID:263310483} {Efficient streaming language models with attention sinks}.
\newblock \emph{ArXiv}, abs/2309.17453.

\bibitem[{Yang et~al.(2024{\natexlab{a}})Yang, Yang, Hui, Zheng, Yu, Zhou, Li, Li, Liu, Huang, Dong, Wei, Lin, Tang, Wang, Yang, Tu, Zhang, Ma, Xu, Zhou, Bai, He, Lin, Dang, Lu, Chen, Yang, Li, Xue, Ni, Zhang, Wang, Peng, Men, Gao, Lin, Wang, Bai, Tan, Zhu, Li, Liu, Ge, Deng, Zhou, Ren, Zhang, Wei, Ren, Fan, Yao, Zhang, Wan, Chu, Cui, Zhang, and Fan}]{Yang2024Qwen2TR}
An~Yang, Baosong Yang, Binyuan Hui, Bo~Zheng, Bowen Yu, Chang Zhou, Chengpeng Li, Chengyuan Li, Dayiheng Liu, Fei Huang, Guanting Dong, Haoran Wei, Huan Lin, Jialong Tang, Jialin Wang, Jian Yang, Jianhong Tu, Jianwei Zhang, Jianxin Ma, Jin Xu, Jingren Zhou, Jinze Bai, Jinzheng He, Junyang Lin, Kai Dang, Keming Lu, Ke-Yang Chen, Kexin Yang, Mei Li, Min Xue, Na~Ni, Pei Zhang, Peng Wang, Ru~Peng, Rui Men, Ruize Gao, Runji Lin, Shijie Wang, Shuai Bai, Sinan Tan, Tianhang Zhu, Tianhao Li, Tianyu Liu, Wenbin Ge, Xiaodong Deng, Xiaohuan Zhou, Xingzhang Ren, Xinyu Zhang, Xipin Wei, Xuancheng Ren, Yang Fan, Yang Yao, Yichang Zhang, Yunyang Wan, Yunfei Chu, Zeyu Cui, Zhenru Zhang, and Zhi-Wei Fan. 2024{\natexlab{a}}.
\newblock \href {https://api.semanticscholar.org/CorpusID:271212307} {Qwen2 technical report}.
\newblock \emph{ArXiv}, abs/2407.10671.

\bibitem[{Yang et~al.(2024{\natexlab{b}})Yang, Han, Gao, Hu, Zhang, and Zhao}]{yang-etal-2024-pyramidinfer}
Dongjie Yang, Xiaodong Han, Yan Gao, Yao Hu, Shilin Zhang, and Hai Zhao. 2024{\natexlab{b}}.
\newblock \href {https://doi.org/10.18653/v1/2024.findings-acl.195} {{P}yramid{I}nfer: Pyramid {KV} cache compression for high-throughput {LLM} inference}.
\newblock In \emph{Findings of the Association for Computational Linguistics ACL 2024}, pages 3258--3270, Bangkok, Thailand and virtual meeting. Association for Computational Linguistics.

\bibitem[{Ye et~al.(2025)Ye, Yin, He, Zhang, Howard, Gao, Durrett, and Chen}]{ye25longproc}
Xi~Ye, Fangcong Yin, Yinghui He, Joie Zhang, Yen Howard, Tianyu Gao, Greg Durrett, and Danqi Chen. 2025.
\newblock Longproc: Benchmarking long-context language models on long procedural generation.
\newblock \emph{arXiv preprint}.

\bibitem[{Zaheer et~al.(2020)Zaheer, Guruganesh, Dubey, Ainslie, Alberti, Onta{\~n}{\'o}n, Pham, Ravula, Wang, Yang, and Ahmed}]{Zaheer2020BigBT}
Manzil Zaheer, Guru Guruganesh, Kumar~Avinava Dubey, Joshua Ainslie, Chris Alberti, Santiago Onta{\~n}{\'o}n, Philip Pham, Anirudh Ravula, Qifan Wang, Li~Yang, and Amr Ahmed. 2020.
\newblock \href {https://api.semanticscholar.org/CorpusID:220831004} {Big bird: Transformers for longer sequences}.
\newblock \emph{ArXiv}, abs/2007.14062.

\bibitem[{Zhang et~al.(2024{\natexlab{a}})Zhang, Chen, Hu, Xu, Chen, Hao, Han, Thai, Wang, Liu, and Sun}]{zhang-etal-2024-bench}
Xinrong Zhang, Yingfa Chen, Shengding Hu, Zihang Xu, Junhao Chen, Moo Hao, Xu~Han, Zhen Thai, Shuo Wang, Zhiyuan Liu, and Maosong Sun. 2024{\natexlab{a}}.
\newblock \href {https://aclanthology.org/2024.acl-long.814} {$\infty${B}ench: Extending long context evaluation beyond 100{K} tokens}.
\newblock In \emph{Proceedings of the 62nd Annual Meeting of the Association for Computational Linguistics (Volume 1: Long Papers)}, pages 15262--15277, Bangkok, Thailand. Association for Computational Linguistics.

\bibitem[{Zhang et~al.(2024{\natexlab{b}})Zhang, Sheng, Zhou, Chen, Zheng, Cai, Song, Tian, R{\'e}, Barrett et~al.}]{zhang2024h2o}
Zhenyu Zhang, Ying Sheng, Tianyi Zhou, Tianlong Chen, Lianmin Zheng, Ruisi Cai, Zhao Song, Yuandong Tian, Christopher R{\'e}, Clark Barrett, et~al. 2024{\natexlab{b}}.
\newblock H2o: Heavy-hitter oracle for efficient generative inference of large language models.
\newblock \emph{Advances in Neural Information Processing Systems}, 36.

\bibitem[{Zhao et~al.(2023)Zhao, Gu, Varma, Luo, chin Huang, Xu, Wright, Shojanazeri, Ott, Shleifer, Desmaison, Balioglu, Nguyen, Chauhan, Hao, and Li}]{Zhao2023PyTorchFE}
Yanli Zhao, Andrew Gu, Rohan Varma, Liangchen Luo, Chien chin Huang, Min Xu, Less Wright, Hamid Shojanazeri, Myle Ott, Sam Shleifer, Alban Desmaison, Can Balioglu, Bernard Nguyen, Geeta Chauhan, Yuchen Hao, and Shen Li. 2023.
\newblock \href {https://api.semanticscholar.org/CorpusID:258297871} {Pytorch fsdp: Experiences on scaling fully sharded data parallel}.
\newblock \emph{Proc. VLDB Endow.}, 16:3848--3860.

\bibitem[{Zhong et~al.(2021)Zhong, Yin, Yu, Zaidi, Mutuma, Jha, Awadallah, Celikyilmaz, Liu, Qiu, and Radev}]{zhong-etal-2021-qmsum}
Ming Zhong, Da~Yin, Tao Yu, Ahmad Zaidi, Mutethia Mutuma, Rahul Jha, Ahmed~Hassan Awadallah, Asli Celikyilmaz, Yang Liu, Xipeng Qiu, and Dragomir Radev. 2021.
\newblock \href {https://doi.org/10.18653/v1/2021.naacl-main.472} {{QMS}um: A new benchmark for query-based multi-domain meeting summarization}.
\newblock In \emph{Proceedings of the 2021 Conference of the North American Chapter of the Association for Computational Linguistics: Human Language Technologies}, pages 5905--5921, Online. Association for Computational Linguistics.

\end{thebibliography}

\appendix

\section{Appendix}
\label{sec:appendix}

\subsection{Implementation details}\label{sec:implementation_details}
\paragraph{Compatibility with Flash Attention} FlashAttention \citep{dao2023flashattention2} substantially improves the efficiency of standard attention computation. It reduces data movements on GPU by directly producing the output for the attention blocks without storing the $O(L^2)$ attention matrix. However, we rely on these attention scores to select the top K tokens during the full attention steps  and construct our partial KV cache $C_p$ (lines 9-10 of Algorithm~\ref{fig:algorithm}). To make our method compatible with Flash Attention, we implement an extra step to re-compute the attention score at the full attention step. As we do not perform full attention at every generation step, this does not introduce significant overhead. For methods that require accessing attention score (e.g. H$_2$O), we apply the same procedure to make them compatible with Flash Attention.

\paragraph{Baseline Settings}
For StreamingLLM, we follow the original paper and maintain a cache with 4 sink tokens and $K$ - 4 recent tokens.
For H$_2$O, we set the heavy hitter size and recent cache size to be $K/2$ each following \cite{zhang2024h2o}. 
For SnapKV, we set the observation window size to 1 and the kernel size to 7 for both \methodN~ and SnapKV following \citet{li2024snapkv}. We apply the same aggregation method (max over all query heads) for SnapKV and H$_{2}$O for the GQA models.

\paragraph{Continued pretraining}
We randomly sample a subset of 200k sequences from the Arxiv split of RedPajama dataset\footnote{\url{https://huggingface.co/datasets/togethercomputer/RedPajama-Data-1T}} and filter out sequences with less than 8192 tokens We train \llama{} for one epoch with a global batch size of 64 and a learning rate of 5e-6. We use 20 warm-up steps and a linear schedule with 0 weight decay. We use the AdamW  Optimizer. We use Fully Sharded Data Parallel \citep{Zhao2023PyTorchFE} and 8-bit optimizer \citep{bitsandbytes} to improve training efficiency. Training is done on 4 H100 80 GB GPUs.

\begin{table}
\begin{center}
\footnotesize
\begin{tabular}{@{}lccc@{}}
\toprule
\textbf{Dataset} & \textbf{\# Example} & \textbf{\# In} & \textbf{\# Out} \\ \midrule 
RULER & 1.3K & 32K & <30 \\ 
QMSum & 100 & 10K & 0.1K \\  
GovReport & 100 & 10K & 0.7K\\ 
NovelSumm & 103 & 100K & 1.0K \\ 
HTML To TSV (0.5K) & 50 & 18K & 0.5K \\ 
HTML To TSV (1K) & 50 & 35K & 1.6K \\ 
HTML To TSV (2K) & 50 & 38K & 4.6K \\ 
Chain of Keys & 100 & 22K & 50 \\
\bottomrule
\end{tabular} 
\end{center}
\caption{Dataset statistics. We report the number of tokens for both the input context and output generation for each dataset, as well as total number of examples.}
\label{tab:dataset_statistics}
\end{table}



\begin{table*}
\setlength{\tabcolsep}{4.5pt}
\begin{center}
\small
\begin{tabular}{@{}lcccccc@{}}
\toprule
& Vanilla & H$_2$O & StreamingLLM & {SnapKV} & \methodN~ (Ours)  \\ 
\midrule 
\textbf{Memory} &   $L$  & $K$ & $K$ & {$K$} & $L + K$\\
\textbf{Time} & $T \times L$ & $T \times K$ & $T \times K$ & {$T \times K$} & $T \times \frac{L}{S}$ + $T \times K$ &  \\ 
\bottomrule
\end{tabular} 
\end{center}
\caption{Comparing memory (KV cache size for $L$ input tokens) and time (attention computation for generating the next $T$ tokens) of \methodN~ and baselines. We denote $S$ as stride and use the same KV cache size ($K$) for the partial cache for our method and complete cache for eviction-based baselines.}
\label{tab:method_comparison}
\end{table*}

\subsection{Memory and time requirement comparison}\label{subsec:memtime}
Table~\ref{tab:method_comparison} compares the memory  and attention compute requirements of \method with baselines. We report the memory required to store the KV cache for the $L$ input tokens, and attention compute required to generate the next $T$ tokens.\footnote{The KV memory requirements also increases with $T$. We do not account for this in the table.} 
We set our partial cache to be the same size as the complete cache of the eviction-based methods. Under this setting, \methodN~requires larger KV cache memory compared to eviction-based baselines, but similar to vanilla attention ($L + K$ vs $L$, where $K << L$). However, our decoding latency is on par with the baselines. Our efficiency depends on two sets of hyperparameters -- the partial cache size $K$, and QC stride and $s$, which determines how often full attention is performed. By setting $K << L$ and a large $S$, we can achieve wall clock times similar to KV eviction-based baselines.

\begin{table}
\begin{center}
\footnotesize
\begin{tabular}{@{}llcc@{}}
\toprule
\textbf{Method}    & \textbf{Agg} & \textbf{Llama-3.1-8B} & \textbf{Qwen-2-7B} \\ 
\midrule
\rowcolor{LightGrey} Vanilla  & - & 2.22/7.07 & 2.33/8.26 \\
RefreshKV   & First &  2.34/7.43 & 2.49/8.78 \\ 
RefreshKV   & Mean & 2.32/7.40 & 2.47/8.73 \\ 
RefreshKV  & Max & {2.32}/{7.40} & 2.47/8.72 \\ 
\bottomrule
\end{tabular} 
\caption{Results comparing different methods to aggregate attention scores for GQA models. We experiment with taking the attention score of the first query head, the average and max attention scores of the query heads in the same group to select topK KV cache. For StreamingLLM and RefreshKV, we set $K=1/8L$ and stride as 10.}
\label{tab:query_aggregation_results}
\end{center}
\end{table}

\subsection{Attention score aggregation for models with GQA}\label{sec:gqa_ablation}

We report language modeling results with different aggregation methods across attention scores of query heads in the same group for models with Grouped Query Attention in Table \ref{tab:query_aggregation_results}. We see that aggregating over the attention score of the entire group works better than using attention score of one of the head, with taking the max slightly outperforming mean.

\subsection{Tuning \textit{s} for query similarity schedule}\label{sec:query_sim}
To choose a similarity threshold $s$ for the dynamic schedule, we run \methodN~on a held-out set of 50 examples from the Book split of the RedPajama dataset. We evaluate on \textit{QC} stride of \{5, 10\} with threshold $s$ of \{0.80, 0.85, 0.90, 0.95\} for \llama{} and \qwen{}.

Table \ref{tab:query_similarity_hyperparameter} reports the results of different settings for perplexity and decoding time measured on one A100 machine with batch size of 1. We can see that for \llama{}, setting a threshold of 0.85 achieves similar performance for both stride compared to 0.90 and 0.95. In contrast the performance of \qwen{} continues to increase going from threshold of 0.80 to 0.95. Therefore, we set the threshold to  0.85 for \llama{} and 0.95 for \qwen{}.

\begin{table}
\begin{center}
\footnotesize
\begin{tabular}{@{}lllcc@{}}
\toprule
\textbf{Method} & \textbf{QC stride} & \textbf{s} & \textbf{Book PPL} & \textbf{Time}\\ 
\midrule
\multicolumn{5}{c}{\textit{Llama-3.1-8B}} \\
\rowcolor{LightGrey} Vanilla & - & - & 6.70 & 7.54 \\
RefreshKV & 5 & 0.80 & 6.92 & 6.42 \\ 
RefreshKV & 5 & 0.85 & \textbf{6.86} & 6.64 \\ 
RefreshKV & 5 & 0.90 & 6.88 & 7.01 \\ 
RefreshKV & 5 & 0.95 & 6.88 & 7.53 \\
RefreshKV & 10 & 0.80 & {6.95} & 6.37 \\ 
RefreshKV & 10 & 0.85 & 6.96 & 6.52 \\ 
RefreshKV & 10 & 0.90 & 6.96 & 6.54 \\ 
RefreshKV & 10 & 0.95 & {6.95} & 7.07 \\
\midrule
\multicolumn{5}{c}{\textit{Qwen-2-7B}} \\
\rowcolor{LightGrey} Vanilla & - & - & 7.44 & 9.11 \\ 
RefreshKV & 5 & 0.80 & 7.86 & 6.50 \\ 
RefreshKV & 5 & 0.85 & 7.80 & 6.64 \\ 
RefreshKV & 5 & 0.90 & 7.73 & 6.91 \\ 
RefreshKV & 5 & 0.95 & \textbf{7.66} & 7.14 \\
RefreshKV & 10 & 0.80 & 7.95 & 6.37 \\ 
RefreshKV & 10 & 0.85 & 7.87 & 6.41 \\ 
RefreshKV & 10 & 0.90 & 7.84 & 6.62 \\
RefreshKV & 10 & 0.95 & 7.82 & 6.67 \\ 
\bottomrule
\end{tabular}
\caption{Results of different similarity threshold $s$ on the held-out set of the Book dataset across two $QC$ stride.}
\label{tab:query_similarity_hyperparameter}
\end{center}
\end{table}

\begin{figure*}[t]
    \centering\vspace{-1em}
    \includegraphics[scale=0.23,trim=0mm 0mm 0mm 0mm]{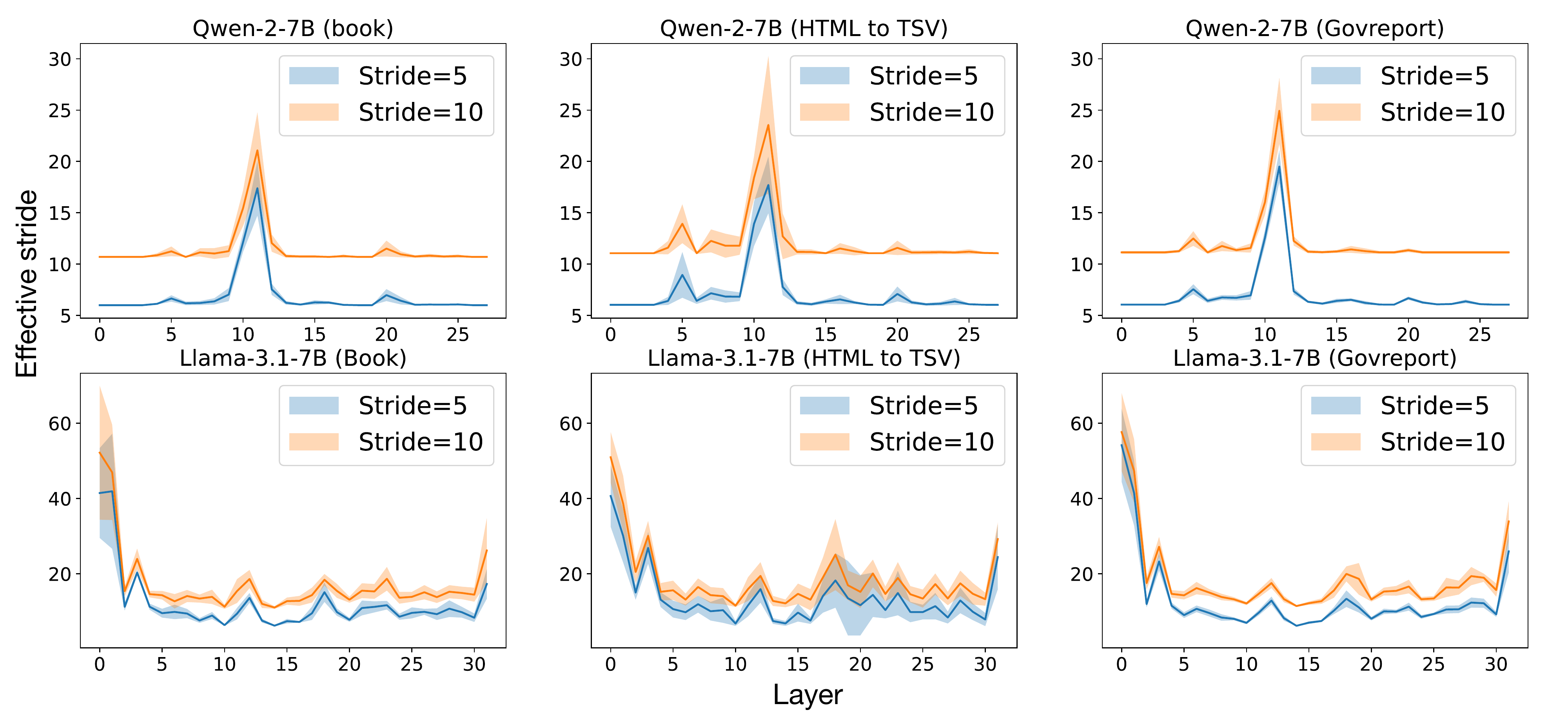}
    \caption{Effective stride across layer for \llama{} (similarity threshold=0.85) and \qwen{} (similarity trheshold=0.95) in three datasets. We sample 10 examples from each dataset to esimate the effective stride.} \vspace{-1em} 
    \label{fig:effective_stride}
\end{figure*}

\subsection{Effective stride}\label{subsec:effective_stride}
We plot the effective stride across layers for \llama{} and \qwen{} in Figure \ref{fig:effective_stride} for the three tasks reported in Table \ref{tab:dynamic_stride}. Leveraging query similarity enables dynamic strides across layers for both models. We observe distinct pattern for the two models, with \llama{} having a larger stride in the first few layer and \qwen{} in the middle layer. We also observe slightly different patterns for different tasks, showing that our method enables flexible scheduling based on the context.

\subsection{Chain-of-key task set-up}\label{subsec:chain_of_keys}
\paragraph{Task set-up} The model is provided with a long lists of keys, each of which contains $W$ number of words, for instance:
\texttt{apricot-waggish} where $W=2$. The model is tasked to generate a sequence which consists of a list of $T$ keys from the context, such that the first word of the next key is the last word of the current key. For example: \texttt{waggish-fishery, fishery-mosquito, mosquito-perfume, perfume-panda, panda-juice} for $T=5$. We provide an example input in Table \ref{tab:chain_of_key_example}.

\paragraph{Data generation} We first generate a list of English words. We then pair each word with another word to form a list of keys. We ensure that for each key $k_{1}$ in the context, there exists exactly one other key 
 $k_{2}$ that satisfies the constraint (i.e. the first word of $k_{2}$ is the last word of $k_{1}$). The keys are randomly shuffled in the context.

\paragraph{Evaluation} We evaluate correctness of the  generated output by the length of a valid chain, divided by $T$. A valid chain needs to satisfy two criteria: (a) all the key must be in the context and (b) the first word of the current key must be the last word of the previous key. We provide example outputs and their correctness score in Table \ref{tab:chain_of_key_eval_example}.

\begin{table}
\begin{center}
\footnotesize
\begin{tabular}{@{}ll|cccc@{}}
\toprule
\textbf{Method} & \textbf{Stride} & \textbf{0.5K} & \textbf{2K} & \textbf{8K} & \textbf{Aggregated} \\ 
\midrule
\multicolumn{6}{c}{\textit{Llama-3.1-8B}} \\
\midrule
\textbf{Vanilla} & & 43 & 31 & 23 & 33 \\
\textbf{Streaming} & & 4 & 1  & 0 & 2 \\ 
\textbf{SnapKV} &  & 0 & 0 & 0 & 0  \\ 
\textbf{H2O} &  & 0 & 0 & 0 & 0 \\
\textbf{Refresh} & QC=5 & \textbf{31} & \textbf{15} & \textbf{4} &  \textbf{17} \\ 
\textbf{Refresh} & QC=10 & 16 & 7 & 1 & 8 \\ 
\midrule 
\multicolumn{6}{c}{\textit{Qwen-2-7B}} \\
\midrule
\textbf{Vanilla} & & 36 & 22 & 15 & 24 \\ 
\textbf{Streaming} & & 10 & \textbf{3} & 0 & 5  \\ 
\textbf{SnapKV} & & 0 & 0 & 0 & 0 \\ 
\textbf{H2O} & &  0 & 0 & 0 & 0 \\
\textbf{Refresh} & QC=5 & \textbf{20} & \textbf{6} & \textbf{3} & \textbf{10} \\
\textbf{Refresh} & QC=10 & 14 & 2 & 1 & 6 \\
\bottomrule
\end{tabular} 
\end{center}
\caption{Breakdown of HTML tasks based on output length.}
\label{tab:html_task_breakdown}
\end{table}

\begin{table*}[ht!]
\small
\begin{tabular}{p{14cm}}
\toprule
\textbf{Input} \\ 
\midrule
“You are given many keys composed of a few words. Your task is to generate a chain of 10 keys such that the first word of the current key is the last word of the previous key. Separate the keys with comma. Example: waggish-fishery, fishery-mosquito, mosquito-perfume, perfume-panda, panda-juice, juice-willow, willow-bronco, bronco-creditor, creditor-bathhouse, bathhouse-woman. You must generate keys that are in the context. DO NOT REPEAT THE EXAMPLE.

Context:Name of key: toga-roommate

Name of key: appetiser-cenario

Name of key: normalization-tacit

Name of key: intensity-ping

Name of key: innate-cummerbund

Name of key: tentacle-lining
[...omitted...] 

Name of key: breath-yielding

Name of key: schema-festive

You are given many keys composed of a few words. Your task is to generate a chain of 10 keys such that the first word of the current key is the last word of the previous key. Separate the keys with comma.You must generate keys that are in the context. Chain of ten keys:” \\
\bottomrule
\end{tabular} 
\caption{Example input for the chain-of-key task where $W=2$ and $T=10$.}
\label{tab:chain_of_key_example}
\end{table*}

\begin{table*}[ht!]
\small
\begin{tabular}{p{10cm}p{3cm}}
\toprule
\textbf{Output} & \textbf{Score}   \\ 
\midrule
 \texttt{impossible-crawdad, crawdad-vehicle, vehicle-uncertainty, uncertainty-welfare, welfare-outrigger, outrigger-historical, historical-gator, gator-hugger, hugger-debris, debris-precious} & 1 (fully correct) \\ \midrule
 \texttt{annoying-pentagon, pentagon-fit, \textcolor{red}{fit-waggish}, waggish-fishery, \textcolor{red}{fishery-mosquito}, mosquito-perfume, perfume-panda, \textcolor{red}{panda-juice}, juice-willow, willow-bronco} & 0.2 (correct up to the second key) \\ \midrule
 \texttt{impossible-crawdad, crawdad-vehicle, vehicle-uncertainty, welfare-outrigger, outrigger-historical, historical-gator, gator-hugger, hugger-debris, debris-precious, uncertainty-welfare} & 0.3 (correct up to the third key) \\
\bottomrule
\end{tabular} \vspace{-0.3em}
\caption{Example output for the chain-of-key task where $W=2$ and $T=10$ and their score. Keys that are not in the context are highlighted in \textcolor{red}{red}.}
\label{tab:chain_of_key_eval_example}
\end{table*}

\subsection{Results on LongProc tasks with short inputs}\label{sec:more_longproc}

\begin{table*}
\footnotesize
\setlength{\tabcolsep}{2.5pt}
\begin{center}
\begin{tabular}{@{}ll|c|c|c|c@{}}
\toprule
\textbf{Method} & \textbf{stride} & \textbf{Path Traversal} &   \textbf{ToM Tracking} & \textbf{Countdown} & \textbf{Travel Planning}\\ 
\midrule
\multicolumn{6}{c}{\textit{Llama-3.1-8B}} \\
\rowcolor{LightGrey} Vanilla & - & 17 & 40 & 67 & 62\\ 
StreamingLLM - & & 0 & 0 & 0 & 0\\
H$_2$O & - & 0 & 0 & 0 & 2  \\ 
SnapKV & - & 1 & 0 & 12 & 0 \\ 
RefreshKV & QC=5 & \textbf{5} & \textbf{14} & \textbf{44} &  \textbf{38} \\
RefreshKV & QC=10 & 1 & 5 & 42 & 18 \\ 
\midrule
\multicolumn{6}{c}{\textit{Qwen-2-7B}} \\
\rowcolor{LightGrey} Vanilla & - & 7 & 12  & 11 &  48 \\ 
StreamingLLM & - & 2 & 0 & 6 & 0 \\
H$_2$O & - & 2 & 0  & 6 & 0 \\ 
SnapKV & - & 0 & 0 &  \textbf{14} & 2  \\ 
RefreshKV & QC=5 & \textbf{3} & \textbf{6} & \textbf{14} & \textbf{26} \\
RefreshKV & QC=10 & 2  & 2 & 10 & 4 \\ 

\bottomrule
\end{tabular} 
\end{center}
\caption{Performance on long-context tasks with short outputs from LongProc benchmark for LLaMA-3.1-8B-Instruct and Qwen-2-7B-Instruct.}
\label{tab:long_proc_results}
\end{table*}

\paragraph{Task set-up} We report results on 4 more tasks from \textbf{LongProc}~\citep{ye25longproc}: \textbf{Path Traversal}, \textbf{Travel Planning}, \textbf{Countdown} and \textbf{Theory-of-mind tracking}. These tasks consist of input with less than 10K tokens. While \textbf{Path Traversal} consists of a version with 12K input tokens, we exclude it from our main results as none of the open sourced models are able to perform the task in vanilla setting. We report results on 50 samples for each task. We set $K=1/8L$ for \methodN~and baselines.

\paragraph{Evaluation} We follow evaluation practice of the original paper~\citep{ye25longproc}. For Countdown and Travel Planning, we report correctness of the final solution using rule-based validators. For Path Traversal and ToM Tracking, we report accuracy.

\paragraph{Results} Results of \methodN~ and baseline methods are in Table \ref{tab:long_proc_results}. We observe similar trend as the \textbf{HTML to TSV} task -- Most of the baselines fail completely on the task. ~\methodN with $QC=5$ recovers ~50\% and ~60\% performance of full attention for \llama{} and \qwen{} respectively.

\subsection{Detailed RULER results}\label{sec:ruler_finegrained}

We follow the suite of evaluation tasks introduced in \cite{hsieh2024ruler}, which consists of the 13 tasks.\footnote{\url{https://github.com/hsiehjackson/RULER}} 
We refer the readers to \citet{hsieh2024ruler} for detailed description and examples of each task and Appendix B for the exact tasks configurations. We group them based on the types:

\begin{itemize}[leftmargin=*]
\item \textbf{Single NIAH} An NIAH-styled task with one key and one value to retrieve. We include three variations of the task with different types of key, value and haystack.
\item \textbf{Multi-key NIAH} An NIAH-styled task with distracting keys. We include three variations of the task with different types of key, value and haystack.
\item \textbf{Multi-value NIAH} An NIAH-styled task with multiple values corresponding to the key. 
\item \textbf{Multi-query NIAH} An NIAH-styled task with multiple queries, each corresponding to a distinct key.
\item \textbf{Variable Tracking} A NIAH-styled task that requires tracing through multiple hops.
\item \textbf{Common word extraction} and \textbf{Frequent word extraction} require extracting the words based on the pattern in a list of words. Common word extraction expects a list of 10 most common words while frequent word extractions expect a list of 3 frequent words.
\item \textbf{Question Answering} A task that requires answering a question given a set of documents. We include two variations of the tasks, corresponding to two question answering datasets.
\end{itemize}

\paragraph{Per-task results} We report detailed performance of RULER subtasks in Table \ref{tab:ruler_detail}, grouped by task type. For both models, the best baselines (SnapKV) achieves comparable results as \methodN~for tasks with short-form outputs, such as \textbf{Single NIAH}. However, for tasks that require longer outputs, such as \textbf{Multi-key} and \textbf{Multi-value NIAH}, \methodN~outperform all the baselines.

\begin{table*}
\setlength{\tabcolsep}{2.5pt}
\small
\begin{center}
\begin{tabular}{@{}l|cccccccc@{}}
\toprule
\textbf{Method}  & \textbf{niah\_single} & \textbf{multi\_key} & \textbf{multi\_query} & \textbf{multi\_value} & \textbf{fwe} & \textbf{vt} & \textbf{cwe} & \textbf{qa} \\
\midrule
\multicolumn{9}{c}{\textit{Llama-3.1-8B}} \\
\rowcolor{LightGrey} {Vanilla}  & 100 &  98 & 99 & 99 & 93 & 99 & 65 & 61  \\
H$_2$O & 7 & 7 & 6 & 6 & 78 & 38 & 39 & 34  \\
{Streaming}  & 8 & 13 & 13 & 13 & 93 & 12 & 4 & 42 \\
SnapKV & 99 & 60 & \textbf{98} & \textbf{99} &\textbf{83} & \textbf{99} & \textbf{44} & \textbf{63} \\
RefreshKV(QC=5) & \textbf{100} & \textbf{91}  & \textbf{98} & \textbf{99} & 81 & \textbf{99} & \textbf{44} & 60\\
RefreshKV(QC=10) & \textbf{100} & 67 & 97 & \textbf{99} & 81 & \textbf{99} & \textbf{44} & 59\\
\midrule
\multicolumn{9}{c}{\textit{Qwen-2-7B}} \\
\rowcolor{LightGrey} {Vanilla} & 100 & 90 & 75 & 87 & 84 & 86 & 27 & 50\\
H$_2$O & 5 & 8 & 5 & 3 & \textbf{84} & 2 & 17 & 30\\ 
{Streaming} & 8 & 11 & 13 & 12 & 80 & 15 & 14 & 39 \\
SnapKV & 69 & 51 & 54 & 43 & {81} & \textbf{87} & \textbf{27} & \textbf{50} \\
RefreshKV(QC=5) & \textbf{99} & \textbf{79} & \textbf{70} & \textbf{85} & 70 & \textbf{87} &  \textbf{27} & \textbf{50} \\
RefreshKV(QC=10) & 97 & 54 & 63 & 67 & 80 & 87 & \textbf{27} & 49\\
\bottomrule
\end{tabular} 
\end{center}
\caption{Detailed performance of RULER subtasks with $L=32K$. For non-vanilla methods, we set the $K=1/8L$.}
\label{tab:ruler_detail}
\end{table*}

\end{document}